\pdfoutput=1
\documentclass[11pt]{article}

\usepackage[]{EMNLP2023}

\usepackage{times}
\usepackage{latexsym}
\usepackage[T1]{fontenc}
\usepackage[utf8]{inputenc}

\usepackage{inconsolata}

\usepackage{hyperref}
\usepackage{url}
\usepackage{svg}
\usepackage{multirow}
\usepackage{algorithm}
\usepackage{algorithmic}

\usepackage{newfloat}
\usepackage{listings}
\floatstyle{ruled}
\newfloat{listing}{tb}{lst}{}
\floatname{listing}{Listing}

\usepackage{graphicx}
\usepackage{amsmath,amssymb,amsthm,mathabx,amsfonts}
\usepackage{bbm}
\usepackage{acronym}
\usepackage{enumitem}
\usepackage{balance}
\usepackage{xspace}
\usepackage{setspace}
\usepackage{url}
\usepackage{amsfonts}
\usepackage{multirow}
\usepackage{booktabs}
\usepackage[switch]{lineno}
\usepackage{multicol,lipsum}
\usepackage{tikz}
\usepackage{pgfplots}
\usepackage{pgfplotstable}
\pgfplotsset{compat=1.17}
\makeatletter
\DeclareRobustCommand\onedot{\futurelet\@let@token\@onedot}
\def\@onedot{\ifx\@let@token.\else.\null\fi\xspace}
 
\def\eg{\emph{e.g}\onedot}

\makeatother

\makeatletter
\newcommand{\thickhline}{%
    \noalign {\ifnum 0=`}\fi \hrule height 1pt
    \futurelet \reserved@a \@xhline
}

\newcommand{\model}{{\fontfamily{lmtt}\selectfont TReE}\xspace}
\title{Enhance Reasoning Ability of Visual-Language Models\\ via Large Language Models}

\author{Yueting Yang$^{1}$, 
Xintong Zhang$^{1}$, 
and Wenjuan Han$1^{1\,*}$ \\
\textsuperscript{1} Beijing Jiaotong University, Beijing, China \\
}

\begin{document}
\maketitle
\begin{abstract}
Pre-trained visual language models (VLM) have shown excellent performance in image caption tasks. However, it sometimes shows insufficient reasoning ability. In contrast, large language models (LLMs) emerge with powerful reasoning capabilities. Therefore, we propose a method called \model, which transfers the reasoning ability of a large language model to a visual language model in zero-shot scenarios. \model contains three stages: observation, thinking, and re-thinking. \textit{Observation} stage indicates that VLM obtains the overall information of the relative image. \textit{Thinking} stage combines the image information and task description as the prompt of the LLM, inference the rationals. \textit{Re-Thinking} stage learns from rationale and then inference the final result through VLM.
% Observe, the first step of understanding is through the visual language model, obtaining the overall information of the image, and generating caption of image. Think,  the image caption information is used as part of prompt, and then LLM generates the reasoning process of the task result. Re-think, the task content and reasoning process are combined as context, and the visual language model is further used to generate task results. 
% 1) In order to verify the enhancement of the model's reasoning ability, we selected a series of visual question-answering datasets for experiments. Compared with the baseline model, our method has significant improvement on the zero-shot visual question-answering task; 2) To verify the enhancement of the model's in-context learning ability, we selected Rendered SST-2 and RavenIQ. The result exceeds the KOSMOS ; 3) Training the model using new datasets generated by our method can also achieve the effect of improving the model's reasoning ability.
\end{abstract}

% introduction figure
% \begin{figure*}[!ht]
% 	\centering
%         \includegraphics[width=\textwidth]{images/figure_1.pdf}
% 	% \includesvg[width=\textwidth]{images/figure_1.svg}
% 	\caption{example }
%         \label{fig:introduction example}
% \end{figure*}

% \begin{figure*}[!ht]
% 	\centering
%         \includegraphics[width=\textwidth]{images/figure_3.pdf}
% 	% \includesvg[width=\textwidth]{images/figure_1.svg}
% 	\caption{example }
%         \label{fig:introduction example}
% \end{figure*}

\begin{figure*}[!ht]
	\centering
        \includegraphics[width=\textwidth]{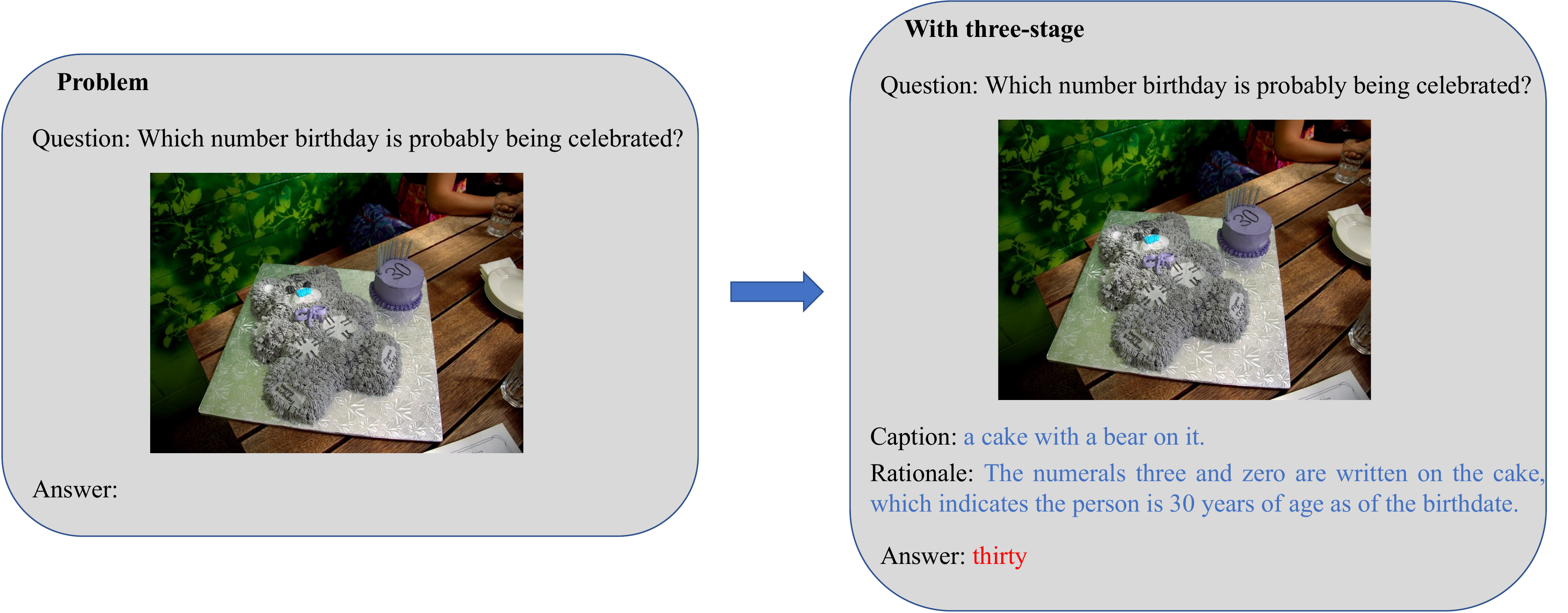}
	% \includesvg[width=\textwidth]{images/figure_1.svg}
	\caption{Illustration of \model.}
        \label{fig:introduction example}
\end{figure*}

\section{Introduction}
% With the development of language models(LMs) in the field of natural language processing(NLP), they have been applied to a variety of tasks, such as common sense reasoning\citep{lan2019albert,brown2020language}, knowledge question answering\citep{shah2019kvqa,talmor2018commonsenseqa}, etc. However, for humans, when performing tasks related to NLP, images can play a very useful role as auxiliary information, instead of relying solely on single-modal information such as text. Therefore, a visual language model(VLM) has been developed\citep{radford2021learning,chen2023vlp,li2022blip,li2023blip}.
Humans interact with the world primarily through vision and language. In recent years, the Vision-language model (VLM) has made significant strides, with multimodal models of increasing scale being developed to push the boundaries of various downstream tasks~\citep{radford2021learning,chen2023vlp,li2022blip,li2023blip}. Building on the success of Large language models (LLMs) in reasoning tasks~\citep{garg2022can,brown2020language}, researchers anticipate that VLMs should also have the ability to process a few training examples and a test instance as its natural language instruction, and directly decode the output without requiring any updates to its parameter. This ability will greatly expand the application prospects of artificial intelligence in both industry and daily life.
% Therefore, it is reasonable to apply LLMs to PLMs in vision and language tasks.

To improve the reasoning ability of VLMs, prompt tuning is the most feed-forward and effective method~\citep{zhou2022learning,zhou2022conditional,rao2022denseclip}. Rooted on the vanilla prompt tuning, \citeauthor{khattak2023maple} propose Multimodal Prompt Learning (MaPLe) for both vision and language branches to improve the alignment between the vision and language representations while \citeauthor{zhang2023prompt} proposes a cascade of foundation models that incorporates diverse prior knowledge. 

Prompt-based approaches coordinate various vision models via LangChain~\citep{langchain}/LLMs, such as Visual ChatGPT~\citep{wu2023visual}, X-GPT~\citep{xia2021xgpt}, MM-REACT~\citep{yang2023mm}.
However, these methods are not only complicated but also cumbersome to implement. Meanwhile, the size of the improved model is usually enlarged, making the model no longer ``simple''.
% However, these approaches have limitations. In order to enhance the performance of VLM, these methods are not only complicated but also cumbersome to implement. At the same time, the size of the optimized model is usually enlarged, so that the model is no longer ``simple''. To mitigate the challenge, a question was raised: How can we improve the model's performance without expanding its size?
% there are usually two types of methods: the first category is to improve the model as a whole and then fine-tune it for different downstream tasks(??); the second category is to improve each part of the model and fine-tune them separately mainly focus on the visual module so that they perform well (prompt tuning ??).
% However, these methods have limitations. To strong the VLM, these methods are both complex and difficult to achieve. After optimized, the model size usually expands, making the model no longer “portable." To mitigate the challenge, a question was raised: How can we improve the model's performance without expanding its size?
% Whether optimizing the model as a whole or only optimizing a certain part, it is often very difficult,

More recently, there is prior work aiming to transfer the LLM's ability to another model. \citeauthor{magister2022teaching} use the LLM as the teacher model and then enhance the student model through knowledge distillation. \citeauthor{liu2022multi} propose a multi-stage prompting approach to generate knowledgeable responses from a single pre-trained LM. Through multi-round conversations, ChatGPT extracts visual information from VLM and summarizes the image content \citep{chen2023video}. \citeauthor{min2021metaicl} adopts meta-learning to enhance the ability of the LM in other tasks through pre-training on different tasks, and achieve the purpose of improving the in-context learning(ICL) ability.

In this paper, we aim to fill this gap and explore a plug-in method without modification for the model architecture and parameters. We propose \model, which is enhanced by the powerful reasoning capabilities of a LLM to assist the VLM (\eg, BLIP-2\citep{li2023blip}) in various downstream tasks.
\model comprises three stages: observation, thinking, and rethinking. During the observation stage, the VLM conducts an initial perception of the image provided by the task, relaying relevant information about the image to the language model. In the thinking stage, the LLM engages in a thorough thought process by producing rationales for various tasks. Finally, the VLM synthesizes the produced rationales in a second thinking stage, which we refer to as rethinking, to accomplish the task. This approach can enhance the VLM's reasoning ability without necessitating any training or fine-tuning.
In summary, our contributions are as follows:
\begin{itemize}
    \item [(1)] We propose a three-stage approach named \model to transfer the reasoning ability of LLM to VLM without any finetuning or new data annotation. We are the first to improve the reasoning ability without the need for finetuning and solely through plug-in LLM;
    \item [(2)] By incorporating LLM as the reasoning processor, PLMs are able to better understand the nature of the question being asked and provide more accurate responses. This has been demonstrated in multiple experiments across various reasoning tasks like RavenIQ dataset~\citep{huang2023language}. Additionally, we have found that our approach is flexible and can be applied to other general visual question-answering task tasks as well, making it a valuable plug-in tool.
    \item [(3)] Moreover, fine-tuning the VLM based on the rationals generated by \model will further improve the reasoning ability. This is also more computationally efficient compared with conventional finetuning methods.  
    % \item[(4)]  \model is more interpretable as it retains the entire reasoning process that leads to the prediction. 
\end{itemize}
% \begin{itemize}
%     \item[(1)] we propose a method that uses the reasoning ability of a large language model to assist a small visual language model to better complete visual question answering tasks (including, VQA, GQA, OK-VQA), and the accuracy rate far exceeds that of models that do not use our method.
%     \item[(2)] With the help of a large-scale language model, the context learning ability of the visual language model (BLIP2) has been improved, and it has completed tasks that could not be completed by means of question and answer, such as Rendered SST-2 and RavenIQ, and obtained on these two data sets. Added new SOTA.
% \end{itemize}

\section{Related Works}
% \subsection{Visual Question Answering task} Visual Question Answering is a task that involves answering questions about an image\citep{agrawal2016vqa}. The goal of general VQA is that model can understand the content of an image and answer questions about it in natural language\citep{hudson2019gqa,talmor2018commonsenseqa}. One of the special VQA tasks like knowledge-based VQA\citep{shah2019kvqa,marino2019okvqa,schwenk2022aokvqa}, requires external knowledge in addition to the image content to answer a question. 
%  \citeauthor{wu2016visual}  divided some of the previous work into four main categories: Joint embedding approaches\citep{gao2015talking,kim2016multimodal}, Attention mechanisms\citep{shih2016look,fukui2016multimodal}, Compositional Models\citep{xiong2016dynamic}, and Models using external knowledge bases\citep{wang2015explicit}. Recently, \citeauthor{yang2022empirical} propose a method named PICa a simple effective method that Prompts GPT3 via the use of Image Captions.
%  The above methods may not be able to introduce reasoning processes into the visual language model. In this study, we propose a method that enhances the reasoning ability of the visual language model by incorporating the reasoning process of VQA obtained by prompting GPT-3 into the visual language model, thereby better completing VQA tasks.
% Previous studies mainly focus on detecting more and more key points about the question (??) or proposed various ways of retrieving and using the multiple knowledge from website resources (??).  \textcolor{red}{todo}
\subsection{Visual Language Model}
Visual language pre-training (VLP) aims to improve the performance of downstream vision and language tasks by pre-training models on large-scale image-text pairs. In order to better complete multi-modal natural language processing tasks, it is necessary to consider unifying vision and language into one framework\citep{cho2021unifying,wang2021simvlm}. This requires designing a model architecture that performs understanding- and generation-based tasks. Existing encoder-based models \citep{radford2021learning} and encoder-decoder models \citep{cho2021unifying,wang2021simvlm} perform suboptimally on the task. And a single unified encoder-decoder \citep{zhou2020unified} limits the model's capability. 

To address the above issues, \citeauthor{li2022blip} propose a multi-modal hybrid encoder-decoder model which provides greater flexibility and better performance on a wide range of downstream tasks while keeping pre-training simple and efficient.
Further considering the end-to-end training of large-scale models, the cost of vision and language pre-training is relatively high. \citeauthor{li2023blip} again proposes a general and effective pre-training strategy ``BLIP-2'' to bootstrap visual-language pre-training from off-the-shelf frozen pre-trained image encoders and frozen large-scale language models. BLIP-2 bridges the modality gap by pre-training a lightweight query converter in two stages, greatly improving training efficiency while saving training costs.

\subsection{In-context learning}
In-context learning(ICL) has become a new paradigm for NLP\citep{garg2022can,brown2020language}. GPT-3 has shown powerful in-context few-shot learning abilities\citep{brown2020language}. Instead of fine-tuning a pre-trained model to adapt it to a downstream task\citep{wei2023larger}, in-context few-shot learners quickly adapt to new tasks with just a few examples in the inference process and require no parameter updates\citep{li2023context}, including question answering, commonsense reasoning, etc. In these tasks, GPT-3 demonstrated a strong reasoning ability to understand the tasks and reason about the results, which means that we can use GPT-3 to reverse the reasoning process of the answer. In our study, we make full use of GPT-3's ICL capabilities to accomplish our goals.

\subsection{Chain of Thought Reasoning}
Chain of Thought(CoT) techniques encourage the LLM to generate intermediate reasoning chains for solving a problem. A reasoning chain is composed of a rationale (a series of intermediate reasoning steps) and an expected answer. 

Previous studies have shown that LLMs can perform CoT reasoning with two major paradigms of techniques: Zero-Shot-CoT and Manual-CoT. Zero-shot-cot, by adding a prompt like “Let’s think step by step” after the test question to invoke CoT reasoning\citep{kojima2022large}. Manual-CoT by eliciting the CoT reasoning ability with effective manual demonstrations\citep{zhou2022least,wang2022self,wang2022rationale}. The demonstrations for the reasoning process are manually designed. Both two paradigms are limited by designing the demonstration manually. Auto-CoT paradigm to automatically construct demonstrations with questions and reasoning chains\citep{zhang2022automatic}. 
In order to expand this method to a visual language model,\citeauthor{zhang2023multimodal} proposed multimodel CoT, through fine-tuning small language models by fusing the vision and language features to perform CoT reasoning. 
In our study, in order to apply CoT to the visual language model more easily, inspired by Auto-CoT, we decided to make full use of the in-context learning ability of large language models and designed a specific prompt paradigm to allow the model to automatically generate rationale, so as to better help VLM to achieve reasoning.
% overview image
% \begin{figure*}[!ht]
% 	\centering
%          \includegraphics[width=\textwidth]{images/figure_2.pdf}
% 	% \includesvg[width=\textwidth]{images/figure_2.svg}
% 	\caption{zero-shot reasoning overview }
%         \label{fig:overview example}
% \end{figure*}
%
\section{\model}

As shown in the figure, our method is mainly divided into three stages. In the first stage \textit{Observation}, the visual language model first understands the image information in the task, by generating the caption of the corresponding image; in the second stage \textit{Thinking}, the LLM according to  the related information based on the task (eg. Caption; Question;) generate reasoning process(Rationale); the third stage \textit{Re-Thinking}, combine the reasoning information from the \textit{Think} stage to  understanding and inference the final result.

\paragraph{\textit{Observation}} The VLM processes the image,  and inference the rough information related to the task for the first time. For example, for the VQA task, the caption of the image is used to assist the reasoning process of the LLM in the second stage.

\paragraph{\textit{Thinking}} LLM has strong in-context learning capabilities. They can understand what tasks are to be completed based on a few simple input-output pairs, and show good reasoning capabilities when completing tasks. Therefore, we consider migrating this ability to a small VLM.
Fully utilize the in-context learning ability of large models, without task description, by using prompt "Question:{} Answer:{[Rationale]. So the answer is}" to generate the corresponding reasoning process(Rationale) Additionally, image is an indispensable and most important piece of information in the VLM. Finally, we update the prompt: "Caption:{} Question:{} Answer:{[Rationale]. So the answer is}" to generate inferences that take image information into account.

\paragraph{\textit{Re-thinking}} After the Reasoning process obtained by LLM, we assume that the VLM understands rationale. When completing the specific task, input rationale as part of context into the VLM and then do \textit{Re-think}. In this part, different prompts are used for different tasks, see the appendix for details.

% As shown in the figure, our method is mainly divided into two stages. In the first stage THINK, the reasoning process is generated by a large language model. In the second stage Re-THINK, the visual language model completes rationale understanding and task inference.
% \paragraph{THINK} Large-scale language models have strong contextual learning capabilities. They can understand what tasks are to be completed based on a few simple input-output pairs, and show good reasoning capabilities when completing tasks. Therefore, we consider migrating this ability to a smaller visual language model, by using "Question:{} Answer:{} Rationale:" to generate the corresponding reasoning process. Image is an indispensable and most important piece of information in the visual language model. We use "Caption:{} Question:{} Answer:{} Rationale:{}" to generate inferences that take image information into account
% \paragraph{Re-THINK} After obtaining the inference process obtained by a large language model, we assume that the visual language model understands rationale. When completing the QA task, input rationale as context into the visual language model for Re-think. In this part, different prompts are used for different tasks, see the appendix.

\begin{table*}[!ht]
\centering
% \usepackage{multirow}
\begin{tabular}{lcccc}
\hline
\multicolumn{1}{c}{Model} & \begin{tabular}[c]{@{}c@{}}VQAv2\\ val\end{tabular} & \begin{tabular}[c]{@{}c@{}}GQA\\ test-dev\end{tabular} & \begin{tabular}[c]{@{}c@{}}OKVQA\\ test\end{tabular} & \begin{tabular}[c]{@{}c@{}}A-OKVQA\\ test\end{tabular} \\ \hline
% \multicolumn{5}{l}{Representative and SoTA methods with numbers reported in the literature}                                                                                                                                                              \\
Flamingo3B                & 49.2                                                & -                                                      & 41.2                                                 & -                                                      \\
Flamingo9B                & 51.8                                                & -                                                      & 44.7                                                 & -                                                      \\
Flamingo80B               & 56.3                                                & -                                                      & 50.6                                                 & -                                                      \\
BLIP2 ViT-L opt2.7B       & 50.1                                                & 33.9                                                   & 30.2                                                 & -                                                      \\
BLIP2 ViT-G opt2.7B       & 53.5                                                & 34.6                                                   & 31.7                                                 & -                                                      \\
BLIP2 ViT-L opt6.7B       & 54.3                                                & 36.4                                                   & 36.4                                                 & -                                                      \\
BLIP2 ViT-L FlanT5XL      & 62.6                                                & 44.4                                                   & 39.4                                                 & -                                                      \\
BLIP2 ViT-G  FlanT5XL     & 63.1                                                & 44.2                                                   & 40.7                                                 & -                                                      \\
BLIP2 ViT-L  FlanT5XXL    & 65.2                                                & 44.7                                                   & 45.9                                                 & -                                                      \\
Unified IO small          & 57.7                                                & -                                                      & 31.0                                                 & 24.3                                                   \\
Unified IO base           & 61.8                                                & -                                                      & 37.8                                                 & 28.5                                                   \\
Uninfied IO large         & 67.8                                                & -                                                      & 42.7                                                 & 33.4                                                   \\
Unified IO xl             & 77.9                                                & -                                                      & 54.0                                                 & 45.2                                                   \\ \hline

% \multicolumn{5}{l}{Our Method  w/ traning}                                                                                                                                                                                                               \\
Ours           &  -                                                   &   -                                                     & -                                                     & \textbf{48.0}                                                       \\ \hline
\end{tabular}
\caption{The results of visual reasoning tasks. The backbone model comes from BLIP2-opt-6.7B\url{https://huggingface.co/Salesforce/blip2-opt-6.7b}.}
\label{tab:main results-reasoning}
\end{table*}

\section{Experiments}
\paragraph{Dataset}
To evaluate how well rationale performs on the initial task of BLIP2\citep{li2023blip}, we conduct experiments on VQAv2\citep{goyal2017making}, OK-VQA\citep{marino2019okvqa}, GQA\citep{hudson2019gqa}, and A-OKVQA\citep{schwenk2022aokvqa}. We also do nonverbal reasoning tasks in RavenIQ dataset at \url{https://aka.ms/kosmos-iq50}. 
% In order to further verify that rationale can improve the in-context learning ability of BLIP2, we selected two datasets: Rendered SST-2\footnote{\url{https://openaipublic.azureedge.net/clip/data/rendered-sst2.tgz}}\citep{radford2021learning} and RavenIQ\footnote{\url{https://aka.ms/kosmos-iq50}}, which are difficult for the visual language model, to conduct experiments. 

% \paragraph{Setup}

\paragraph{Baselines}
\begin{itemize}
    \item We chose BLIP2 models of different sizes as the baseline for comparison, including BLIP2 model at \url{https://github.com/salesforce/LAVIS/tree/main/lavis/models/blip2_models}; 
    \item To compare the performance in general, we choose Flanmingo\citep{alayrac2022flamingo} and Unified-IO\citep{lu2022unified} to compare with.
    \item In order to compare the performance of our method in improving BLIP2 in-context learning ability, we selected a large multi-modal language model KOSMOS\citep{huang2023language} for comparison on the new task.
\end{itemize}

\section{Results}
\subsection{General Vision-Language Reasoning Tasks}
Our main research is to transfer the reasoning ability of the GPT-3.5 \citep{brown2020language} to BLIP2, so as to enhance the reasoning ability of BLIP2 on the question-answering task and the in-context learning ability on the new task. Therefore, we mainly consider the experimental comparison from two perspectives: zero-shot reasoning ability and in-context learning ability. We evaluate these tasks by accuracy.

\subsection{Nonverbal Reasoning Task}
From the experimental results in Table \ref{tab:ravenIQ result}, it can be seen that the visual language model using the model method outperforms KOSMOS on the RavenIQ task

\begin{table}[!ht]
\centering
\begin{tabular}{lc}
\hline
Method                  & RavenIQ(\%) \\ \hline
Random                  & 17          \\
KOSMOS-1                & 22          \\ \hline
% Our Method w/o training &             \\
Ours         & 27          \\ \hline
\end{tabular}
\caption{Results of RavenIQ.}
\label{tab:ravenIQ result}
\end{table}

\section{Conclusion}
Despite the compelling boosted results, the performance on all tasks is far from satisfactory. Nevertheless, this work sheds light on enhancing multimodal ICL ability and calls for future research in this direction.

\bibliography{anthology,custom}

\begin{thebibliography}{37}
\expandafter\ifx\csname natexlab\endcsname\relax\def\natexlab#1{#1}\fi

\bibitem[{Alayrac et~al.(2022)Alayrac, Donahue, Luc, Miech, Barr, Hasson, Lenc,
  Mensch, Millican, Reynolds et~al.}]{alayrac2022flamingo}
Jean-Baptiste Alayrac, Jeff Donahue, Pauline Luc, Antoine Miech, Iain Barr,
  Yana Hasson, Karel Lenc, Arthur Mensch, Katherine Millican, Malcolm Reynolds,
  et~al. 2022.
\newblock Flamingo: a visual language model for few-shot learning.
\newblock \emph{Advances in Neural Information Processing Systems},
  35:23716--23736.

\bibitem[{Brown et~al.(2020)Brown, Mann, Ryder, Subbiah, Kaplan, Dhariwal,
  Neelakantan, Shyam, Sastry, Askell et~al.}]{brown2020language}
Tom Brown, Benjamin Mann, Nick Ryder, Melanie Subbiah, Jared~D Kaplan, Prafulla
  Dhariwal, Arvind Neelakantan, Pranav Shyam, Girish Sastry, Amanda Askell,
  et~al. 2020.
\newblock Language models are few-shot learners.
\newblock \emph{Advances in neural information processing systems},
  33:1877--1901.

\bibitem[{Chen et~al.(2023{\natexlab{a}})Chen, Zhang, Han, Chen, Shi, Xu, and
  Xu}]{chen2023vlp}
Fei-Long Chen, Du-Zhen Zhang, Ming-Lun Han, Xiu-Yi Chen, Jing Shi, Shuang Xu,
  and Bo~Xu. 2023{\natexlab{a}}.
\newblock Vlp: A survey on vision-language pre-training.
\newblock \emph{Machine Intelligence Research}, 20(1):38--56.

\bibitem[{Chen et~al.(2023{\natexlab{b}})Chen, Zhu, Haydarov, Li, and
  Elhoseiny}]{chen2023video}
Jun Chen, Deyao Zhu, Kilichbek Haydarov, Xiang Li, and Mohamed Elhoseiny.
  2023{\natexlab{b}}.
\newblock \href {http://arxiv.org/abs/2304.04227} {Video chatcaptioner: Towards
  enriched spatiotemporal descriptions}.

\bibitem[{Cho et~al.(2021)Cho, Lei, Tan, and Bansal}]{cho2021unifying}
Jaemin Cho, Jie Lei, Hao Tan, and Mohit Bansal. 2021.
\newblock Unifying vision-and-language tasks via text generation.
\newblock In \emph{International Conference on Machine Learning}, pages
  1931--1942. PMLR.

\bibitem[{Garg et~al.(2022)Garg, Tsipras, Liang, and Valiant}]{garg2022can}
Shivam Garg, Dimitris Tsipras, Percy~S Liang, and Gregory Valiant. 2022.
\newblock What can transformers learn in-context? a case study of simple
  function classes.
\newblock \emph{Advances in Neural Information Processing Systems},
  35:30583--30598.

\bibitem[{Goyal et~al.(2017)Goyal, Khot, Summers-Stay, Batra, and
  Parikh}]{goyal2017making}
Yash Goyal, Tejas Khot, Douglas Summers-Stay, Dhruv Batra, and Devi Parikh.
  2017.
\newblock \href {http://arxiv.org/abs/1612.00837} {Making the v in vqa matter:
  Elevating the role of image understanding in visual question answering}.

\bibitem[{Huang et~al.(2023)Huang, Dong, Wang, Hao, Singhal, Ma, Lv, Cui,
  Mohammed, Liu et~al.}]{huang2023language}
Shaohan Huang, Li~Dong, Wenhui Wang, Yaru Hao, Saksham Singhal, Shuming Ma,
  Tengchao Lv, Lei Cui, Owais~Khan Mohammed, Qiang Liu, et~al. 2023.
\newblock Language is not all you need: Aligning perception with language
  models.
\newblock \emph{arXiv preprint arXiv:2302.14045}.

\bibitem[{Hudson and Manning(2019)}]{hudson2019gqa}
Drew~A. Hudson and Christopher~D. Manning. 2019.
\newblock \href {http://arxiv.org/abs/1902.09506} {Gqa: A new dataset for
  real-world visual reasoning and compositional question answering}.

\bibitem[{Khattak et~al.(2023)Khattak, Rasheed, Maaz, Khan, and
  Khan}]{khattak2023maple}
Muhammad~Uzair Khattak, Hanoona Rasheed, Muhammad Maaz, Salman Khan, and
  Fahad~Shahbaz Khan. 2023.
\newblock \href {http://arxiv.org/abs/2210.03117} {Maple: Multi-modal prompt
  learning}.

\bibitem[{Kojima et~al.(2022)Kojima, Gu, Reid, Matsuo, and
  Iwasawa}]{kojima2022large}
Takeshi Kojima, Shixiang~Shane Gu, Machel Reid, Yutaka Matsuo, and Yusuke
  Iwasawa. 2022.
\newblock Large language models are zero-shot reasoners.
\newblock \emph{arXiv preprint arXiv:2205.11916}.

\bibitem[{Langchain(2022)}]{langchain}
Langchain. 2022.
\newblock Langchain, \url{https://github.com/hwchase17/langchain}.
\newblock \emph{Langchain}.

\bibitem[{Li et~al.(2023{\natexlab{a}})Li, Li, Savarese, and Hoi}]{li2023blip}
Junnan Li, Dongxu Li, Silvio Savarese, and Steven Hoi. 2023{\natexlab{a}}.
\newblock Blip-2: Bootstrapping language-image pre-training with frozen image
  encoders and large language models.
\newblock \emph{arXiv preprint arXiv:2301.12597}.

\bibitem[{Li et~al.(2022)Li, Li, Xiong, and Hoi}]{li2022blip}
Junnan Li, Dongxu Li, Caiming Xiong, and Steven Hoi. 2022.
\newblock Blip: Bootstrapping language-image pre-training for unified
  vision-language understanding and generation.
\newblock In \emph{International Conference on Machine Learning}, pages
  12888--12900. PMLR.

\bibitem[{Li et~al.(2023{\natexlab{b}})Li, Gong, Feng, Xu, Zhang, Wu, and
  Kong}]{li2023context}
Mukai Li, Shansan Gong, Jiangtao Feng, Yiheng Xu, Jun Zhang, Zhiyong Wu, and
  Lingpeng Kong. 2023{\natexlab{b}}.
\newblock In-context learning with many demonstration examples.
\newblock \emph{arXiv preprint arXiv:2302.04931}.

\bibitem[{Liu et~al.(2022)Liu, Patwary, Prenger, Prabhumoye, Ping, Shoeybi, and
  Catanzaro}]{liu2022multi}
Zihan Liu, Mostofa Patwary, Ryan Prenger, Shrimai Prabhumoye, Wei Ping,
  Mohammad Shoeybi, and Bryan Catanzaro. 2022.
\newblock Multi-stage prompting for knowledgeable dialogue generation.
\newblock \emph{arXiv preprint arXiv:2203.08745}.

\bibitem[{Lu et~al.(2022)Lu, Clark, Zellers, Mottaghi, and
  Kembhavi}]{lu2022unified}
Jiasen Lu, Christopher Clark, Rowan Zellers, Roozbeh Mottaghi, and Aniruddha
  Kembhavi. 2022.
\newblock Unified-io: A unified model for vision, language, and multi-modal
  tasks.
\newblock \emph{arXiv preprint arXiv:2206.08916}.

\bibitem[{Magister et~al.(2022)Magister, Mallinson, Adamek, Malmi, and
  Severyn}]{magister2022teaching}
Lucie~Charlotte Magister, Jonathan Mallinson, Jakub Adamek, Eric Malmi, and
  Aliaksei Severyn. 2022.
\newblock \href {http://arxiv.org/abs/2212.08410} {Teaching small language
  models to reason}.

\bibitem[{Marino et~al.(2019)Marino, Rastegari, Farhadi, and
  Mottaghi}]{marino2019okvqa}
Kenneth Marino, Mohammad Rastegari, Ali Farhadi, and Roozbeh Mottaghi. 2019.
\newblock \href {http://arxiv.org/abs/1906.00067} {Ok-vqa: A visual question
  answering benchmark requiring external knowledge}.

\bibitem[{Min et~al.(2021)Min, Lewis, Zettlemoyer, and
  Hajishirzi}]{min2021metaicl}
Sewon Min, Mike Lewis, Luke Zettlemoyer, and Hannaneh Hajishirzi. 2021.
\newblock Metaicl: Learning to learn in context.
\newblock \emph{arXiv preprint arXiv:2110.15943}.

\bibitem[{Radford et~al.(2021)Radford, Kim, Hallacy, Ramesh, Goh, Agarwal,
  Sastry, Askell, Mishkin, Clark et~al.}]{radford2021learning}
Alec Radford, Jong~Wook Kim, Chris Hallacy, Aditya Ramesh, Gabriel Goh,
  Sandhini Agarwal, Girish Sastry, Amanda Askell, Pamela Mishkin, Jack Clark,
  et~al. 2021.
\newblock Learning transferable visual models from natural language
  supervision.
\newblock In \emph{International conference on machine learning}, pages
  8748--8763. PMLR.

\bibitem[{Rao et~al.(2022)Rao, Zhao, Chen, Tang, Zhu, Huang, Zhou, and
  Lu}]{rao2022denseclip}
Yongming Rao, Wenliang Zhao, Guangyi Chen, Yansong Tang, Zheng Zhu, Guan Huang,
  Jie Zhou, and Jiwen Lu. 2022.
\newblock \href {http://arxiv.org/abs/2112.01518} {Denseclip: Language-guided
  dense prediction with context-aware prompting}.

\bibitem[{Schwenk et~al.(2022)Schwenk, Khandelwal, Clark, Marino, and
  Mottaghi}]{schwenk2022aokvqa}
Dustin Schwenk, Apoorv Khandelwal, Christopher Clark, Kenneth Marino, and
  Roozbeh Mottaghi. 2022.
\newblock \href {http://arxiv.org/abs/2206.01718} {A-okvqa: A benchmark for
  visual question answering using world knowledge}.

\bibitem[{Wang et~al.(2022{\natexlab{a}})Wang, Wei, Schuurmans, Le, Chi, and
  Zhou}]{wang2022rationale}
Xuezhi Wang, Jason Wei, Dale Schuurmans, Quoc Le, Ed~Chi, and Denny Zhou.
  2022{\natexlab{a}}.
\newblock Rationale-augmented ensembles in language models.
\newblock \emph{arXiv preprint arXiv:2207.00747}.

\bibitem[{Wang et~al.(2022{\natexlab{b}})Wang, Wei, Schuurmans, Le, Chi, and
  Zhou}]{wang2022self}
Xuezhi Wang, Jason Wei, Dale Schuurmans, Quoc Le, Ed~Chi, and Denny Zhou.
  2022{\natexlab{b}}.
\newblock Self-consistency improves chain of thought reasoning in language
  models.
\newblock \emph{arXiv preprint arXiv:2203.11171}.

\bibitem[{Wang et~al.(2021)Wang, Yu, Yu, Dai, Tsvetkov, and
  Cao}]{wang2021simvlm}
Zirui Wang, Jiahui Yu, Adams~Wei Yu, Zihang Dai, Yulia Tsvetkov, and Yuan Cao.
  2021.
\newblock Simvlm: Simple visual language model pretraining with weak
  supervision.
\newblock \emph{arXiv preprint arXiv:2108.10904}.

\bibitem[{Wei et~al.(2023)Wei, Wei, Tay, Tran, Webson, Lu, Chen, Liu, Huang,
  Zhou et~al.}]{wei2023larger}
Jerry Wei, Jason Wei, Yi~Tay, Dustin Tran, Albert Webson, Yifeng Lu, Xinyun
  Chen, Hanxiao Liu, Da~Huang, Denny Zhou, et~al. 2023.
\newblock Larger language models do in-context learning differently.
\newblock \emph{arXiv preprint arXiv:2303.03846}.

\bibitem[{Wu et~al.(2023)Wu, Yin, Qi, Wang, Tang, and Duan}]{wu2023visual}
Chenfei Wu, Shengming Yin, Weizhen Qi, Xiaodong Wang, Zecheng Tang, and Nan
  Duan. 2023.
\newblock Visual chatgpt: Talking, drawing and editing with visual foundation
  models.
\newblock \emph{arXiv preprint arXiv:2303.04671}.

\bibitem[{Xia et~al.(2021)Xia, Huang, Duan, Zhang, Ji, Sui, Cui, Bharti, and
  Zhou}]{xia2021xgpt}
Qiaolin Xia, Haoyang Huang, Nan Duan, Dongdong Zhang, Lei Ji, Zhifang Sui,
  Edward Cui, Taroon Bharti, and Ming Zhou. 2021.
\newblock Xgpt: Cross-modal generative pre-training for image captioning.
\newblock In \emph{Natural Language Processing and Chinese Computing: 10th CCF
  International Conference, NLPCC 2021, Qingdao, China, October 13--17, 2021,
  Proceedings, Part I 10}, pages 786--797. Springer.

\bibitem[{Yang et~al.(2023)Yang, Li, Wang, Lin, Azarnasab, Ahmed, Liu, Liu,
  Zeng, and Wang}]{yang2023mm}
Zhengyuan Yang, Linjie Li, Jianfeng Wang, Kevin Lin, Ehsan Azarnasab, Faisal
  Ahmed, Zicheng Liu, Ce~Liu, Michael Zeng, and Lijuan Wang. 2023.
\newblock Mm-react: Prompting chatgpt for multimodal reasoning and action.
\newblock \emph{arXiv preprint arXiv:2303.11381}.

\bibitem[{Zhang et~al.(2023{\natexlab{a}})Zhang, Hu, Li, Huang, Deng, Li, Qiao,
  and Gao}]{zhang2023prompt}
Renrui Zhang, Xiangfei Hu, Bohao Li, Siyuan Huang, Hanqiu Deng, Hongsheng Li,
  Yu~Qiao, and Peng Gao. 2023{\natexlab{a}}.
\newblock Prompt, generate, then cache: Cascade of foundation models makes
  strong few-shot learners.
\newblock \emph{arXiv preprint arXiv:2303.02151}.

\bibitem[{Zhang et~al.(2022)Zhang, Zhang, Li, and Smola}]{zhang2022automatic}
Zhuosheng Zhang, Aston Zhang, Mu~Li, and Alex Smola. 2022.
\newblock Automatic chain of thought prompting in large language models.
\newblock \emph{arXiv preprint arXiv:2210.03493}.

\bibitem[{Zhang et~al.(2023{\natexlab{b}})Zhang, Zhang, Li, Zhao, Karypis, and
  Smola}]{zhang2023multimodal}
Zhuosheng Zhang, Aston Zhang, Mu~Li, Hai Zhao, George Karypis, and Alex Smola.
  2023{\natexlab{b}}.
\newblock \href {http://arxiv.org/abs/2302.00923} {Multimodal chain-of-thought
  reasoning in language models}.

\bibitem[{Zhou et~al.(2022{\natexlab{a}})Zhou, Sch{\"a}rli, Hou, Wei, Scales,
  Wang, Schuurmans, Bousquet, Le, and Chi}]{zhou2022least}
Denny Zhou, Nathanael Sch{\"a}rli, Le~Hou, Jason Wei, Nathan Scales, Xuezhi
  Wang, Dale Schuurmans, Olivier Bousquet, Quoc Le, and Ed~Chi.
  2022{\natexlab{a}}.
\newblock Least-to-most prompting enables complex reasoning in large language
  models.
\newblock \emph{arXiv preprint arXiv:2205.10625}.

\bibitem[{Zhou et~al.(2022{\natexlab{b}})Zhou, Yang, Loy, and
  Liu}]{zhou2022conditional}
Kaiyang Zhou, Jingkang Yang, Chen~Change Loy, and Ziwei Liu.
  2022{\natexlab{b}}.
\newblock \href {http://arxiv.org/abs/2203.05557} {Conditional prompt learning
  for vision-language models}.

\bibitem[{Zhou et~al.(2022{\natexlab{c}})Zhou, Yang, Loy, and
  Liu}]{zhou2022learning}
Kaiyang Zhou, Jingkang Yang, Chen~Change Loy, and Ziwei Liu.
  2022{\natexlab{c}}.
\newblock Learning to prompt for vision-language models.
\newblock \emph{International Journal of Computer Vision}, 130(9):2337--2348.

\bibitem[{Zhou et~al.(2020)Zhou, Palangi, Zhang, Hu, Corso, and
  Gao}]{zhou2020unified}
Luowei Zhou, Hamid Palangi, Lei Zhang, Houdong Hu, Jason Corso, and Jianfeng
  Gao. 2020.
\newblock Unified vision-language pre-training for image captioning and vqa.
\newblock In \emph{Proceedings of the AAAI conference on artificial
  intelligence}, volume 34-07, pages 13041--13049.

\end{thebibliography}
\bibliographystyle{acl_natbib}

\appendix

\section{Prompt Design}
\label{sec:appendix A}
By utilizing the ICL capability of GPT-3 without adding any task related descriptions and using only input-output pairs, we have designed a simple and useful prompt template for different tasks. Table \ref{tab:prompt template} shows more details.

% \usepackage{multirow}
\begin{table*}[!ht]
\centering
\begin{tabular}{l|l|l}
\hline
\multicolumn{1}{c|}{Task Type} & \multicolumn{1}{c|}{"Thinking"  Template}                                                                                                                                                                                          & \multicolumn{1}{c}{"Re-Thinking" Template}                                                                                                               \\ \hline
Question-Answerting            & \begin{tabular}[c]{@{}l@{}}Caption:\{image caption\}\\ Question:\{question\}\\ Answer:\{rationale\}.So the answer  is \{answer\}\end{tabular}                                                                                      & \multirow{3}{*}{\begin{tabular}[c]{@{}l@{}}Question:\{\}\textbackslash{}t\\ Rationale:\{\}\textbackslash{}t\\ Answer:\{\}\textbackslash{}t\end{tabular}} \\ \cline{1-2}
% Rendere SST2                     & \begin{tabular}[c]{@{}l@{}}Sentence:\{image caption\}\\ Answer:\{rationale\},thus the sentence's sentiment is \{answer\}\end{tabular}                                                                                              &                                                                                                                                                          \\ \cline{1-2}
RavenIQ                        & \begin{tabular}[c]{@{}l@{}}The first picture is \{\}.\textbackslash{}n ... The third picture is \{\}.\textbackslash{}n\\ Question: What does the next image look like?\textbackslash{}n\\ Answer:The next pictire is \{\}.\end{tabular} &                                                                                                                                                          \\ \hline
\end{tabular}
\caption{the prompt template in different tasks.}
\label{tab:prompt template}
\end{table*}

\end{document}